\pdfoutput=1

\documentclass[11pt]{article}

\usepackage[]{EMNLP2023}

\usepackage{times}
\usepackage{bigfoot}
\usepackage{latexsym}

\usepackage[T1]{fontenc}

\usepackage[utf8]{inputenc}
\usepackage[ukrainian,english]{babel}

\usepackage{microtype}
\usepackage{inconsolata}
\usepackage{multirow}
\usepackage[spacesep,definevectors]{easyvector}
\usepackage{amsmath}
\usepackage{mathtools}
\usepackage{xspace}
\usepackage{algpseudocode}
\usepackage{algorithm}
\usepackage{hyperref}

\usepackage{inconsolata}

\DeclareMathOperator*{\argmin}{arg\,min}
\DeclareNewFootnote{ANote}[fnsymbol]
 
\newcommand{\OurMODEL}{GRI} 

\newcommand{\eat}[1]{}
\DeclareUnicodeCharacter{0301}{\hspace{-1ex}\'{ }}

\title{\OurMODEL: Graph-based Relative Isomorphism of Word Embedding Spaces}

\author{
    Muhammad Asif Ali,\textsuperscript{\rm 1} 
    Yan Hu,\textsuperscript{\rm 1}
    Jianbin Qin,\textsuperscript{\rm 2}
    Di Wang\textsuperscript{\rm 1}\\
    \textsuperscript{\rm 1} King Abdullah University of Science and Technology, KSA\\
    \textsuperscript{\rm 2} Shenzhen University, China\\
    \{muhammadasif.ali, yan.hu, di.wang\}@kaust.edu.sa, qinjianbin@szu.edu.cn\\
}

\begin{document}
\maketitle
\begin{abstract}
Automated construction of bilingual dictionaries using monolingual 
embedding spaces is a core challenge in machine translation. The end 
performance of these dictionaries relies upon the geometric similarity 
of individual spaces, i.e., their degree of isomorphism. Existing 
attempts aimed at controlling the relative isomorphism of different 
spaces fail to incorporate the impact of semantically related words 
in the training objective. To address this, we propose \OurMODEL~that 
combines the distributional training objectives with attentive graph 
convolutions to unanimously consider the impact of semantically similar 
words required to define/compute the relative isomorphism of multiple spaces. 
Experimental evaluation shows that \OurMODEL~outperforms the existing 
research by improving the average P$@$1 by a relative score of up to 63.6\%.
We release the codes for \OurMODEL~at \url{https://github.com/asif6827/GRI}.
\end{abstract}

\section{Introduction}
\vspace{-1.3ex}
Bilingual Lexical Induction (BLI) aims at the construction of 
lexical dictionaries using different mono-lingual word embeddings.
Automated construction of bilingual dictionaries plays a significant 
role, especially for resource-constrained languages where hand-crafted 
dictionaries are almost non-existent.
It is also a key tool to bootstrap the performance of 
many down-streaming applications, e.g., cross-lingual information 
retrieval~\citep{2018_artetxe}, neural machine translation~
\cite{2018_lample}.

The most prevalent way for the construction of cross-lingual 
embeddings is to map the mono-lingual embeddings in a shared 
space using linear and/or non-linear transformations, also known 
as mapping-based methods~\cite{2017_muse,2018_joulin,2019_patra}.
A core limitation of the mapping-based methods is their reliance
on the approximate isomorphism assumption, i.e., the underlying 
monolingual embedding spaces are geometrically similar. 
This severely limits the applicability of the mapping-based 
methods to closely related languages and similar data domains.
This isomorphism assumption does not hold, especially in case 
of domain-mismatch and for languages exhibiting different 
characteristics~\cite{2017_muse,2018_sogaard,2019_glavas,2019_patra}.
Other dominant factors identified in the literature that limit 
the end performance of BLI systems include: (i) linguistic 
differences (ii) algorithmic mismatch, (iii) variation in 
data size, (iv) parameterization etc. 
Similar to the supervised models, the unsupervised variants of BLI 
are also unable to cater to the above-mentioned challenges~
\cite{2020_kim_unsupervise,2020_marie_iter}.
\begin{figure}[t]
	\centering
	\includegraphics[width=0.85\linewidth]{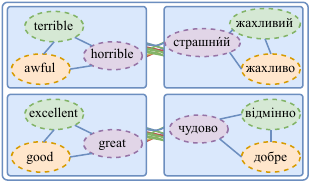}
	\caption{Semantically related tokens for English and Ukrainian languages. These words though lexically varying 
		carry the same semantics and their impact should be unanimously considered.}
	\vspace{-3.7ex}
	\label{fig:isomorphism}
\end{figure}
Instead of relying on embedding spaces trained completely independent 
of each other, in the recent past there have been a shift in explicitly 
using the isomorphism measures alongside distributional training
objective~\cite{2022_isovec}.
In order to control the relative isomorphism of monolingual embedding 
spaces, these models use existing bilingual dictionaries as training 
seeds. However, one core limitation of these models is their inability 
to incorporate the impact of semantically relevant tokens into the 
training objective. This severely deteriorates the relative isomorphism 
of the resultant cross-lingual embedding space.

This phenomenon is illustrated in Figure~\ref{fig:isomorphism}
for English and Ukrainian languages. For example, for the English 
language, we often use terms \{\emph{``terrible", ``horrible"}\} 
within the same context
without a significant change in the meaning of the sentence.
For these terms, corresponding terms in the Ukrainian language
~\{\textcyrillic{``страшни́й''}, \textcyrillic{``жахливо''}\} 
may also be used interchangeably without a significant change in the context.
Likewise, for the bottom row in Figure~\ref{fig:isomorphism},
the words~\{\emph{``good", ``great", ``excellent"}\} are 
semantically related words in the English language, with 
\{\textcyrillic{``відмінно''}, \textcyrillic{``чудово''}, \textcyrillic{``добре''}\}
as corresponding semantically related words in the 
Ukrainian language.

To address these challenges, in this paper we propose a novel
framework named: Graph-based Relative Isomorphism (\OurMODEL).
\OurMODEL~uses attentive graph convolutions to pay attention to 
semantically related tokens, followed by using 
isomorphism metrics to inject this information into the 
model training. Later, it combines the isomorphism loss with the 
distributional training objective to train the complete model.

We argue~\OurMODEL~offers a better alternative for BLI, as it 
allows injecting information about the semantic variations of tokens 
in the training objective, which is a more natural setting in order to control the 
relative isomorphism of linguistic data. 
An immediate benefit of the proposed model is obvious 
in the domain-mismatch settings, where attentive graph convolutions 
mechanism by~\OurMODEL~offer the ability to unanimously
analyze and/or model similar concepts represented by 
lexically varying terms across different corpora.
This is also evident by a relatively stable performance
of \OurMODEL~for both domain-sharing and domain-mismatch
settings (Section~\ref{dis:dm}).
We summarize the core contributions of this paper as follows:

\begin{enumerate}
	\item We propose~\OurMODEL~that combines isomorphism loss functions 
	(guided by graph convolutions) along with the distributional 
	training objective for BLI.
	
	\item We propose attentive graph convolutions for \OurMODEL~in order 
	to control the relative isomorphism by sharing information across 
	semantically related tokens.
	
	\item We illustrate the effectiveness of \OurMODEL~via comprehensive 
	experimentation. For benchmark data sets, \OurMODEL~ outperforms the
	existing state of the art by approximately~63.6\% for average P$@$1. 
\end{enumerate}

\eat{
	The most similar work to ours is probably the
	one of Wada and Iwata (2018), where the authors train a LSTM (Hochreiter and Schmidhuber,
	1997) language model with sentences from different languages. They share the LSTM parameters but use different lookup tables to represent
	the words in each language. They focus on aligning word representations and show that their approach work well on word translation tasks.
}

\eat{
	As an example, 
	In addition, 
	which makes it a natural choice for 
	which makes it more robust to
	it provides the provision to augment the performance of...
	to learn isomorphism 
	Experimentation shows the flexibility of the...~\OurMODEL~ 
	to...}

\vspace{-1.7ex}
\section{Related Work}
\vspace{-0.7ex}
Due to limited space, we primarily categorize the related 
work on relative isomorphism of cross-lingual embeddings 
into: (i) mapping to shared space, and (ii) joint training.
\eat{
	(iii) embeddings \& geometry.
}
\paragraph{Mapping to shared space.}
These models aim to find a linear and/or non-linear transformation 
for pre-trained 
word embeddings in order to map them to a shared space.
These models rely on the assumption that the embedding models
share similar structure across different languages~\cite{2013_mikolov_exploiting}, 
which allows them to independently train embeddings for different 
languages and learn mapping functions to align them in a shared 
space.
Supervised variants in this regard use existing bilingual resources, 
such as parallel dictionaries~\cite{2015_xing, 2018_joulin,2019_jawan}. 
The unsupervised variants use distributional matching~\cite{2017_zhang, 2017_muse, 
	2018_artetxe_robust, 2019_zhou}.
These models have also been applied to the contextualized embeddings
~\cite{2019_aldarmaki,2019_schuster}.

\eat{
	It includes supervised variants as well as unsyVaria~\citep{2016_artetxe, 2017_artetxe, 2018_doval, }.}

\paragraph{Joint Training}
These models put additional constraints on model learning, i.e., a hard or
soft cross-lingual constraints in addition to the monolingual training 
objectives. Similar to the mapping-based models, early works in this 
domain include the supervised variants relying on bilingual 
dictionaries~ \citep{2016_ammar, 2015_luong, 2015_gouws}.
Recently, the unsupervised approaches have gained attention because of 
their ease of implementation. For instance,~\citet{2018_lample}
analyzed the performance for concatenated monolingual corpora with shared
vocabulary without any additional cross-lingual resources. 
Results show that this setting outperforms many carefully crafted alignment
based strategies for unsupervised machine translation.
Other unsupervised approaches with good results on benchmark data 
sets include zero-shot cross-lingual transfer by~\citet{2019_artetxe_massive} 
and cross-lingual pre-training by~\citet{2019_lample_cross}.
\citet{2022_isovec} proposed IsoVec that introduces multiple 
different losses along with the skip-gram loss function to control 
the relative isomorphism of mono-lingual spaces.

A major limitation of these methods is their inability to 
incorporate the lexico-semantic variations of word pairs 
across different languages in the model training, which 
severely limits the end performance of these models.

\eat{A detailed analysis for mapping vs joint training 
	is performed by~\citet{2019_wang}
	
	\paragraph{Embedding \& Geometry}
	These methods analyze the impact of comparable 
	corpora to yield isomorphic embeddings and the role 
	of geometric properties of the embedding spaces.
}

\section{Background}
In this section, we discuss the notation and mathematical 
background of the tools and techniques used in this paper. 

\subsection{Notation}
Throughout this paper, we use $\mathbf{U} \in \mathbf{R}^{p \times d}$ and 
$\mathbf{V} \in \mathbf{R}^{q \times d}$ to represent
the embeddings of the source and target languages. 
We assume the availability of seeds pairs for
both source and target languages, represented by: 
$\{(u_0,v_0), (u_1,v_1),...(u_s,v_s)\}$.

\subsection{VecMap toolkit}
\label{vecmap_bkgd}
For mapping across different embedding spaces, we use vecmap toolkit\footnote{\url{https://github.com/artetxem/vecmap}}.
We follow~\citet{2019_zhang_girls} to pre-process the embeddings, 
i.e., the embeddings are unit-normed, mean-centered and unit-normed 
again. For bilingual induction, we follow the steps outlined by
~\citep{2018_artetxe}, i.e., whitening each space, and solving Procrustes. 
This is followed by re-weighting, de-whitening, and mapping of 
translation pairs via nearest-neighbor retrieval. For details, 
refer to the original work by~\citet{2018_artetxe}.

\section{Proposed Approach}
\subsection{Problem Definition}
In this paper, we address a core challenge in controlling the 
relative isomorphism for cross-lingual data sets, i.e., incorporate 
the impact of semantically coherent words for BLI.

\subsection{Overview}
\begin{figure}[t]
	\centering
	\includegraphics[width=1.05\columnwidth]{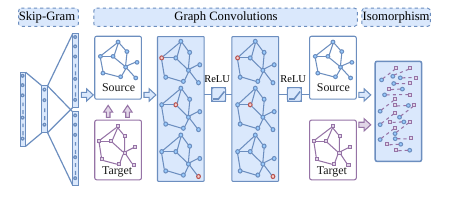}
	\vspace{-3.7ex}
	\caption{Proposed framework for Graph-based Relative Isomorphism(\OurMODEL). 
		It combines attentive graph convolutions with the skip-gram to control the 
		relative isomorphism for source $\mathbf{U}$ and target $\mathbf{V}$ embeddings.}
	\vspace{-3.7ex}
	\label{fig:framework}
\end{figure}

We propose Graph-based Relative Isomorphism \OurMODEL, shown in Figure~\ref{fig:framework},
that aims to learn distributional information in 
the source embedding space $\mathbf{U}$, in such a way:
(i) $\mathbf{U}$ is geometrically similar to the target embedding 
space $\mathbf{V}$ to the best possible extent, 
(ii) $\mathbf{U}$ captures information about the semantically related terms in $\mathbf{V}$.
In order to capture the distributional information 
\OurMODEL~uses skip-gram with negative sampling.
In order to control the geometry and isomorphism of 
embedding space $\mathbf{U}$ relative to space 
$\mathbf{V}$, \OurMODEL~uses attentive graph convolutions. 
Finally, it uses multiple different isomorphism metrics 
along with the skip-gram loss function for model training.

We claim the proposed model provides the provision to perform 
BLI in a performance-enhanced fashion by using attentive 
graph convolutions for effective propagation of semantic 
relatedness of tokens across different languages.

\subsection{\OurMODEL}
\vspace{-1ex}
In order to learn the distributional embeddings for the 
source language that are geometrically similar to the target 
embeddings \OurMODEL~incorporates attentive graph
convolutions along with the distributional training objective.
\OurMODEL~relies on the assumption that each language possesses 
multiple variations of semantically similar tokens that may be 
used interchangeably.
And, in order to effectively model the relative isomorphism 
for the multi-lingual data sets this phenomenon needs to be 
captured explicitly.

The proposed model~(\OurMODEL)~is based on the assumption
that sharing information among semantically related words
is inevitable in order to control the relative isomorphism
of the embedding spaces. 
From the linguistic perspective, there are an arbitrary 
number of words semantically related to a given word, 
which makes graphs a natural choice to unanimously consider 
the impact of these words in the end model.
We explain the individual components of \OurMODEL~in the 
following sub-sections:

\subsubsection{Distributional Representation}
In order to learn the distributional representations 
\OurMODEL~uses skip-gram with negative sampling.
For training the skip-gram model, we follow the same settings 
as outlined by~\citet{2013_mikolov_sg}, i.e., embed a word 
close to its contextual neighbors and far from a set of words 
randomly chosen from the vocabulary. The formulation of the 
skip-gram loss function is illustrated in Equation~\ref{SG_eq}.

\begin{equation}
\label{SG_eq}
\begin{aligned}
\vspace{-2.7ex}
\mathcal{L}_{SG} = \log \sigma({u^{'}_{c_O}}^{\mathsf{T}} u_{c_{I}}) + \\
\sum_{i}^{k} \mathbf{E}_{c_i \sim P_{n}(c)} \big[ \log \sigma (-{u^{'}_{c_i}}^{\mathsf{T}} u_{c_{I}})\big]
\vspace{-2.7ex}
\end{aligned}
\end{equation}
Here $u_{c_O}$ and $u_{c_I}$ correspond to the output and input vector
representation of the word $c$. $u^{'}_{c_i}$ the embedding vectors 
for noise terms. $P_{n}(c)$ corresponds to the noise distribution, 
$k$ is the number of noisy samples. We use $k=10$ in our case.

\subsubsection{Capturing Semantics}
In order to control the relative isomorphism across the source 
and target embeddings~\OurMODEL~uses attentive graph convolutions under 
transductive settings in order to share information among semantically 
related words. The graph construction is summarized in Algorithm~\ref{alg:graphs}, 
and explained as follows:

\paragraph{Graph Construction.}
Inputs for the graph construction include: 
(i) the supervision seed pairs for the target language, 
(ii) existing pre-trained word2vec embeddings\footnote{\url{https://code.google.com/archive/p/word2vec/}, 
	trained using Google-News Corpus of 100 billion words.}~\cite{2013_mikolov_sg}.
The graph construction process proceeds as follows:

Firstly, we organize the target words into all possible pairs, i.e., combinations 
of two words at a time. For each word pair, we compute a score (cosine similarity) 
of their corresponding embedding vectors. The word pairs with scores higher than 
a threshold ($thr$) are stored as the probable semantically related terms ($\text{Pairs}_{prob}$), 
illustrated in lines (2-6). 
We observed that using a significantly higher value for $thr$ is beneficial, because: 
(i) it helps in capturing the most confident word pairs thus overcoming noise, and 
(ii) only a few word pairs end up in $\text{Pairs}_{prob}$ which makes it computationally efficient.
Finally, for all word pairs in $\text{Pairs}_{prob}$, we 
formulate edges to construct the graph $\text{G}$. 
For each word pair, we use the cosine score of corresponding embedding vectors as the 
attention weight. 

\eat{
	for the target language, 
	and their corresponding embeddings from pre-existing open 
	source resources that are used as inputs to our graph 
	construction process.
	In order to preserve the lexical variations across semantically
	relevant tokens, we propose attentive graph convolutions as part 
	of \OurMODEL.
	We exploit the fact that the graph provides a 
	provision to surround each word in the 
	For these words, we formulate all possible pairs.
	For eachof these pairs, we use the corresponding embeddings to 
	compute their similarity.
	Later, the word pairs with similarity surpassing a threshold $thr$
	are considered as probable candidates for the graph construction process.
	For consider all pair 
	For our experiments, we use a relatively higher value of threshold $(thr)$.
	It offers following benefits: (i), (ii) and 
	This phenomenon needs to be captured explicitly
	And, in order to capture/compute the relative isomorphism for 
	the multi-lingual data sets
	This needs to be captured while modeling the joint distribution of
	multi-lingual data set.
	This is illustrated in Figure~\ref{fig:isomorphism}.}

\begin{algorithm}[t]
	\caption{Graph Construction}
	\label{alg:graphs}
	\textbf{Input:} {Embedding (EMB); \\
		$\hspace*{11mm}$ D$_{tar} = \text{Target} (D_{tr + dev + tst})$}\\
	\textbf{Output:} {Graph: $\text{G}$}
	\begin{algorithmic}[1]
		\State{$\text{Pairs}_{prob}$ $\gets$ {$\mathbf{0}$};  $\text{G} \gets \emptyset$}
		\For{$(w_1,w_2) \gets \text{Pairs}(\text{D}_{tar})$}
		\State{$y^{*}$ = $\text{score}_{\text{EMB}}(w_1,w_2)$}
		\If{$y^{*} \geq {thr}$}
		\State{$\text{Pairs}_{prob} \gets \text{Pairs}_{prob} \cup (w_1, w_2)$}
		\EndIf
		\EndFor
		
		\For{$pair \in \text{Pairs}_{prob}$}
		\State{$\text{G}$ $\gets$ $\text{G} \cup \{edge(pair)$\}}
		\EndFor
		\State \Return {$\text{G}$}
		
	\end{algorithmic}
\end{algorithm}

\paragraph{Attentive Graph Convolutions.}
Depending upon the value of $thr$, graph G surrounds each word by a set of highly confident semantically 
related words (including their lexical variations). The degree of similarity is controlled by the cosine 
similarity of embedding vectors. Later, for each word, we aggregate the information contained in the 
neighbors to come up with a new representation of the word that accommodates information from semantically 
related neighbors.

Note, in our setting, unlike the existing work by~\citet{2016_kipf}, we propose attentive graph 
convolutions with pair-wise distributional similarity scores as the hard attention weights. 
The attention weights are not updated during the model training.
Specifically, we use the following layer-wise propagation mechanism:
\begin{equation}
\label{eq:RR}
\vspace{-1.7ex}
L^{(i+1)} = \rho(\Tilde{\Gamma} L^{(i)} W_i)
\end{equation}
where 
$\Tilde{\Gamma}= \bar{D}^{-1/2} (\Gamma + I) \bar{D}^{-1/2}$ is the normalized symmetric matrix, 
$\bar{D}$ is the degree matrix of $\Gamma$, 
$\Gamma$ is the weighted adjacency matrix learned from graph $\text{G}$ with pair-wise scores as the attention weights, 
$L^{(i)}$ is the input from previous layer, with $L^{(0)} = \UU \in \mathbf{R}^{p\times{d}}$ as the 
input matrix corresponding to the embeddings of the source terms, 
$W_{i}$ is the learn-able weight matrix, 
$\rho$ is the non-linear activation function.

Note, the end goal of the attentive convolutions is to analyze each word in $\text{G}$ in relation 
to the weighted combination of its semantic neighbors. 
For this, we surround each word (node) with a group of semantically related words 
(nodes in the graph) and perform weighted aggregation to re-compute the representation 
of the word. We also allow self-connections for each word, i.e., adding an identity 
matrix $I$ to $\Gamma$.
This will enforce "semantically related words" to get similar representations.
We observed, that for our problem settings, this attentive graph convolution 
framework outperforms the basic settings with equal contribution from all 
neighbors~\cite{2016_kipf}.

For \OurMODEL, we use a two-layered network to learn 
the final embeddings of each word ${\UU_{m}} \in \mathbf{R}^{p \times {d}}$ as follows:
\begin{equation}
\vspace{-1.7ex}
\UU_{m} = \Tilde{(\Gamma)}(ReLU(\Tilde{(\Gamma)}\UU W_0)) W_1
\end{equation}
\subsection{Isomorphism Loss functions}
In order to train the \OurMODEL, we experiment with multiple different isomorphism 
loss functions on top of the attentive graph convolution network. 
Details about each loss function are as follows:

\paragraph{L2 loss.}
We use L2-norm averaged over the number of data samples.
\begin{equation}
\vspace{-1.7ex}
\mathcal{L}_{2} = \frac{1}{N} {||\mathbf{U}_{m}-\mathbf{V}||_{2}}
\end{equation}
\paragraph{Orthognal Procrustus loss.}
Orthogonal Procrustes loss aims to find a linear transformation
$W$ to solve:

\begin{equation}
\label{proc_eq}
\mathcal{L}_{proc} = \argmin_{\WW \in \mathbf{R}^{d \times d}, \WW^{T}\WW=I} \frac{1}{N}{||\mathbf{U}_{m}\mathbf{W}-\mathbf{V}||}_{2}
\end{equation}

\citet{1966_schonemann} proposed a solution $\mathbf{W} = \QQ\PP^{T}$, 
where $\PP \Sigma \QQ^{T}$ is the singular value decomposition 
of $\VV^{T}\UU_{m}$.

\paragraph{Procrustus loss with initialization.}
It follows the same process as that of the Procrustus
loss with the exception that we initialize the 
embeddings for the source words with the embedding 
vectors of their translation vectors corresponding 
to the target words. The end goal of this setting is 
to analyze the ability of the \OurMODEL~to propagate 
the knowledge of the initialized embeddings during 
the model training. We also allow updating the initialized 
word embeddings during model training. We denote this loss 
by $\mathcal{L}_{proc_{init}}$. We use the symbol $\mathcal{L}_{ISO}$ to represent the 
different variations of isomorphism losses, i.e., 
$\mathcal{L}_{2}$, $\mathcal{L}_{proc}$ and $\mathcal{L}_{proc_{init}}$.

\subsection{The Complete Model}
Finally, we combine the distributional training objective 
with the isomorphism loss function to compute complete
loss of \OurMODEL, as follows:

\begin{equation}
\label{Eq:Final}
\vspace{-0.7ex}
\mathcal{L}_{\OurMODEL} = \alpha  \mathcal{L}_{SG} + (1-\alpha)  \mathcal{L}_{ISO}
\end{equation}
where $\alpha$ is the parameter used to control the 
contribution $\mathcal{L}_{SG}$ and $\mathcal{L}_{ISO}$ respectively.

\section{Experiments and Results}

\subsection{Datasets}
In order to set up a unanimous platform for comparative analysis, 
we use the data settings used by~\citet{2022_isovec}.
We use the first 1 million lines from newscrawl-2020 data set 
for English (``en"), Bengali (``bn") and Tamil (``ta") and the 
entire of newscrawl-2020 data for Ukrainian (``uk") to train word 
embeddings. We used Moses scripts for data pre-processing\footnote{\href{http://github.com/moses-smt/mosesdecoder/tree/master/scripts/tokenizer}{Moses script}}.
For evaluation, we used publically available train, dev, and test 
splits provided by MUSE~\citep{2017_muse}. 
Out of approx 8000-word pairs for each language, we used
word pairs 0-5000, 5001-6500, and 6501-8000 as train, 
test and dev set respectively.
The train set is used for model training, dev set is 
used for parameter tuning, and final results are 
reported using the test set. All these data splits 
are non-overlapping.

\subsection{Baseline Models}
For comparative evaluation, we use independently trained
distributional embeddings for the source and target 
languages as a baseline model.
Amongst the existing research, we compare \OurMODEL~against 
the prevalent state-of-the-art work on BLI, 
i.e., IsoVec by~\citet{2022_isovec}. IsoVec uses 
the skip-gram training objective along with 
isomorphism training objectives. 
Note,~\citet{2022_isovec} used exactly the same data 
settings as that of our proposed model (i.e., \OurMODEL), 
so for performance comparison, we use the numbers 
reported in the original paper.

\subsection{Experimental Settings}
For model training, we use Adam optimizer~
\citep{2014_adam} with learning rate = 0.001;
$\alpha$ in Equation~\ref{Eq:Final} is set to 0.7;
the value of $thr$ in algorithm~\ref{alg:graphs} is 
set to 0.5. For experiments, we use embeddings learnt for 
English language as the target embeddings, and embeddings
for other languages, i.e., ``ta", ``uk", and ``bn", 
as the source embeddings. For mapping 
across different spaces, we use Vecmap toolkit with 
process-flow explained in Section~\ref{vecmap_bkgd}.
We use average P@1 as the evaluation metric. 
We report mean $(\mu)$ and standard deviation $(\sigma)$ 
of the results over 5 runs of the experiment.
All experiments are performed using Intel 
Core-i9-10900X CPU, and Nvidia 3090Ti GPU.

\subsection{Main Results}

\begin{table}[t]
	\centering
	\resizebox{1.02\columnwidth}{!}{
		\begin{tabular}{lll|clc|lcl}
			\hline
			\multicolumn{3}{l}{Methodology}              & \multicolumn{2}{|c}{bn}                & \multicolumn{2}{|c}{uk}             & \multicolumn{2}{|c}{ta}         \\
			\hline
			\multicolumn{3}{l}{Baseline}                & \multicolumn{2}{|c}{13.1 ($\pm$ 0.51)}    & \multicolumn{2}{|c}{13.9 ($\pm$ 0.45)}  & \multicolumn{2}{|c}{10.8 ($\pm$ 0.42)} \\
			\hline
			\multicolumn{3}{l}{IsoVec (L2)}             & \multicolumn{2}{|c}{16.3 ($\pm$ 0.4)}     & \multicolumn{2}{|c}{16.5 ($\pm$ 0.4)}  & \multicolumn{2}{|c}{11.1 ($\pm$ 0.5)} \\
			\multicolumn{3}{l}{IsoVec (Proc-L2)}        & \multicolumn{2}{|c}{16.6 ($\pm$ 0.7)}     & \multicolumn{2}{|c}{16.0 ($\pm$ 0.8)}  & \multicolumn{2}{|c}{10.7 ($\pm$ 0.3)} \\
			\multicolumn{3}{l}{IsoVec (Proc-L2-Init)}   & \multicolumn{2}{|c}{\underline{16.9} ($\pm$ 0.2)} & \multicolumn{2}{|c}{\underline{17.1} ($\pm$ 0.6)}  & \multicolumn{2}{|c}{\underline{11.8} ($\pm$ 0.3)} \\
			\hline
			\multicolumn{3}{l}{\OurMODEL~($\mathcal{L}_2$)}                 & \multicolumn{2}{|c}{17.28 ($\pm$ 0.02)}   & \multicolumn{2}{|c}{18.75 ($\pm$ 0.41)} & \multicolumn{2}{|c}{13.47 ($\pm$ 0.04)}     \\
			\multicolumn{3}{l}{\OurMODEL~($\mathcal{L}_{{proc}_{init}}$)}   & \multicolumn{2}{|c}{19.83 ($\pm$ 0.05)}   & \multicolumn{2}{|c}{21.37 ($\pm$ 0.08)} & \multicolumn{2}{|c}{15.27 ($\pm$ 0.01)}    \\
			\multicolumn{3}{l}{\OurMODEL~($\mathcal{L}_{proc}$)}            & \multicolumn{2}{|c}{\textbf{20.52} ($\pm$ 0.02)} & \multicolumn{2}{|c}{\textbf{27.97} ($\pm$ 2.63)} & \multicolumn{2}{|c}{\textbf{18.97} ($\pm$ 0.2)} \\
			\hline
	\end{tabular}}
	\caption{\OurMODEL~results for the proposed model. We compare results with IsoVec~\citep{2022_isovec}.}
	\vspace{-3.7ex}
	\label{tab:res1}
\end{table}

The results for the proposed model (\OurMODEL) compared 
with the baseline models are shown in Table~\ref{tab:res1}.
We boldface the overall best-performing scores with the previous 
state-of-the-art underlined.

These results show the~\OurMODEL~has a relatively stable performance 
(with low variance), it consistently outperforms the baseline and 
previous state-of-the-art scores by a significant margin.
For ``bn", ``uk", and ``ta",~\OurMODEL~outperforms the IsoVec~
\citep{2022_isovec} by 21.4\%, 63.6\% and 60.7\% respectively.
Especially noteworthy is the performance improvement gained by
\OurMODEL~for the Ukrainian language.
We attribute this performance improvement to the fact that
the semantic relatedness of the words corresponding to 
the Ukrainian embedding space is relatively better compared 
to other languages.

The performance comparison of different isomorphism loss functions shows 
that $\mathcal{L}_{proc}$ consistently outperforms the $\mathcal{L}_{proc_{init}}$ 
and $\mathcal{L}_2$ across all data sets. A relatively low performance 
of $\mathcal{L}_{proc_{init}}$ compared to the $\mathcal{L}_{proc}$ may be 
attributed to the fact that randomly initialized embeddings are a better 
choice compared to the initialization from the seed pairs. The initialization 
from the seed pairs may not be helpful for the model training to improve 
the performance at later stages.

Overall results show the significance of using attentive 
graph convolutions in controlling the relative geometry 
of source language for BLI.
Especially, the ability of the attentive convolutions 
to accumulate the contribution of semantically related
terms plays a vital role in controlling 
the relative geometry of the source embeddings relative to the target embeddings, 
as is evident from the results in Table~\ref{tab:res1}.

\section{Discussion}

In this sub-section, we perform a detailed analysis of
the performance of~\OurMODEL. Specifically, we analyze:
(i) Domain mis-match settings
(ii) Impact of attentive convolutions, 
(iii) Isometric metrics, and 
(iv) Error cases.

\subsection{Domain mismatch}
\label{dis:dm}
Domain-mismatch has been identified as one of the core
limitations of existing BLI methods. These methods fail
badly in inferring bilingual information for embeddings
trained on data sets from different domains
~\cite{2018_sogaard,2020_marchisio}.

We claim that incorporating lexical variations for 
semantically related tokens makes~\OurMODEL~robust 
to the domain mismatch settings.
In order to validate these claims for \OurMODEL, 
we re-run the experiments using target embeddings 
trained on 33.8 million lines of web-crawl data from the 
English Common Crawl data. The embeddings for the source
languages (``bn", ``uk" and ``ta") are trained 
using the newscrawl-2020 data.
The results for the domain-mismatch experiments for
different isomorphism loss functions are reported 
in Table~\ref{tab:domain}. 

These results are compared against the baseline 
distributional embeddings and best-performing scores of the existing work, i.e., IsoVec by~\citet{2022_isovec}. 
Note, for the domain mismatch experiments, we use 
exactly same data settings as that of~\citet{2022_isovec}, so we report exactly the same numbers as reported in 
original paper.
Comparing the results of our model against the IsoVec, 
the \OurMODEL~improves the performance by 27.74\%, 
53.12\% and 74.22\% for the {``bn"}, {``uk"} and 
{``ta"} languages respectively.

\begin{table}[b]
	\centering
	\resizebox{1.02\columnwidth}{!}{
		\begin{tabular}{lll|clc|lcl}
			\hline
			\multicolumn{3}{l}{Methodology}  & \multicolumn{2}{|c}{bn}  & \multicolumn{2}{|c}{uk}  & \multicolumn{2}{|c}{ta}         \\
			\hline
			\multicolumn{3}{l}{Baseline}     & \multicolumn{2}{|c}{9.7 ($\pm$ 0.72)}   & \multicolumn{2}{|c}{10.2 ($\pm$ 0.43)}   & \multicolumn{2}{|c}{7.5 ($\pm$ 0.39)}          \\
			\hline
			\multicolumn{3}{l}{IsoVec (Proc-L2-Init)} & \multicolumn{2}{|c}{15.5 ($\pm$ 0.7)}  & \multicolumn{2}{|c}{17.3 ($\pm$ 0.4)}  & \multicolumn{2}{|c}{10.9 ($\pm$ 0.5)} \\
			\hline
			\multicolumn{3}{l}{\OurMODEL~($\mathcal{L}_2$)}  & \multicolumn{2}{|c}{13.97 ($\pm$ 0.02)}    & \multicolumn{2}{|c}{17.32 ($\pm$ 0.32)} & \multicolumn{2}{|c}{11.93 ($\pm$ 0.01)}     \\
			\multicolumn{3}{l}{\OurMODEL~($\mathcal{L}_{{proc}_{init}}$)}   & \multicolumn{2}{|c}{19.75 ($\pm$ 0.01)}     & \multicolumn{2}{|c}{21.32 ($\pm$ 0.10)} & \multicolumn{2}{|c}{17.12 ($\pm$ 0.59)}    \\
			\multicolumn{3}{l}{\OurMODEL~($\mathcal{L}_{proc}$)} & \multicolumn{2}{|c}{\textbf{19.80} ($\pm$ 0.50)}  & \multicolumn{2}{|c}{\textbf{26.49} ($\pm$ 0.50)} & \multicolumn{2}{|c}{\textbf{18.99} ($\pm$ 0.20)}     \\
			\hline
	\end{tabular}}
	\caption{\OurMODEL~results for domain mis-match experiments compared with the baseline models, IsoVec~\cite{2022_isovec}.}
	\label{tab:domain}
\end{table}

Comparing these results against the main experiments 
reported in Table~\ref{tab:res1}, we can see 
the \OurMODEL~yields a stable performance for both 
domain-shared as well as domain mismatch settings.
These results show that the attentive graph convolutions
indeed allow information sharing across semantically related tokens along with their 
lexical variations that is in turn helpful in controlling the 
relative isomorphism of the embedding spaces.

Comparing the results for different loss functions, we can see that similar to the main experiments the 
performance of the model for the Procrustes loss $(\mathcal{L}_{proc})$ is relatively higher
than the $\mathcal{L}_2$ and $\mathcal{L}_{proc_{init}}$.

\begin{table}[t]
	\centering
	\resizebox{1.02\columnwidth}{!}{
		\begin{tabular}{lll|clc|lcl}
			\hline
			\multicolumn{3}{l}{Methodology}                & \multicolumn{2}{|c}{bn}  & \multicolumn{2}{|c}{uk}              & \multicolumn{2}{|c}{ta}         \\
			\hline
			\multicolumn{3}{l}{\OurMODEL~w/o G-Conv ($\mathcal{L}_2$)}            & \multicolumn{2}{|c}{16.10 ($\pm$ 0.35)} & \multicolumn{2}{|c}{16.35 ($\pm$ 0.30)}  & \multicolumn{2}{|c}{11.25 ($\pm$ 0.45)} \\
			\multicolumn{3}{l}{\OurMODEL~w/o G-Conv ($\mathcal{L}_{proc_{init}}$)}& \multicolumn{2}{|c}{16.75 ($\pm$ 0.20)} & \multicolumn{2}{|c}{16.98 ($\pm$ 0.30)}  & \multicolumn{2}{|c}{11.70 ($\pm$ 0.25)} \\
			\multicolumn{3}{l}{\OurMODEL~w/o G-Conv ($\mathcal{L}_{proc}$)}       & \multicolumn{2}{|c}{16.50 ($\pm$ 0.5)} & \multicolumn{2}{|c}{16.10 ($\pm$ 0.70)}  & \multicolumn{2}{|c}{10.65 ($\pm$ 0.20)} \\
			\hline
			\multicolumn{3}{l}{\OurMODEL~($\mathcal{L}_2$)}                 & \multicolumn{2}{|c}{17.28 ($\pm$ 0.02)}   & \multicolumn{2}{|c}{18.75 ($\pm$ 0.41)} & \multicolumn{2}{|c}{13.47 ($\pm$ 0.04)}     \\
			\multicolumn{3}{l}{\OurMODEL~($\mathcal{L}_{{proc}_{init}}$)}   & \multicolumn{2}{|c}{19.83 ($\pm$ 0.05)}   & \multicolumn{2}{|c}{21.37 ($\pm$ 0.08)} & \multicolumn{2}{|c}{15.27 ($\pm$ 0.01)}    \\
			\multicolumn{3}{l}{\OurMODEL~($\mathcal{L}_{proc}$)}            & \multicolumn{2}{|c}{\textbf{20.52} ($\pm$ 0.02)} & \multicolumn{2}{|c}{\textbf{27.97} ($\pm$ 2.63)} & \multicolumn{2}{|c}{\textbf{18.97} ($\pm$ 0.20)} \\
			\hline
	\end{tabular}}
	\caption{Analyzing the impact of attentive graph convolutions for \OurMODEL.}
	\vspace{-3.7ex}
	\label{tab:gc}
\end{table}

\subsection{Impact of attentive convolutions}
\label{dis:att_conv}
In this sub-section, we analyze in detail the performance 
improvement of~\OurMODEL~attributable to the attentive 
graph convolutions. 
For this, we primarily analyze the performance 
improvement of~\OurMODEL~with and without attentive 
graph convolutions. The results of these experiments are 
reported in Table~\ref{tab:gc}. These results show the 
significance of attentive graph convolutions that help 
in improving the performance across all three languages.
The improvement in performance for the ``bn", ``uk" and 
``ta" languages is 24.36\%, 64.72\% and 62.13\% 
respectively.

To gain further insight, we also analyzed the output of
the model with and without graph convolutions in order 
to dig out which class of translation instances were correctly 
translated especially due to the attentive convolutions 
part of~\OurMODEL. We run this analysis only for the Ukrainian 
language because: \OurMODEL~ yields a higher score for the Ukrainian 
language compared to other languages. All the analyses were
performed under the direct supervision of a linguistic expert.

Detailed analyses show that a major portion (approx 51\%) of 
the pairs corrected especially by the graph convolutions 
belong to the nouns, with 21\% verbs and 20\% adjectives.
The rest 7\% are assigned to other classes. This analysis 
shows that the phenomenon of lexical variation is dominant 
among nouns that results in better performance of~\OurMODEL~
compared to the baseline models. 

\subsection{Isometric metrics}
We also correlate the results of~\OurMODEL~with different widely 
used isomorphism metrics. Specifically, we use two metrics, namely: 
(a) Pearson's correlation, and (b) Eigenvector similarity.
Details about these metrics and the corresponding experimental 
setting are as follows:

\begin{table}[t]
	\centering
	\resizebox{0.92\columnwidth}{!}{
		\begin{tabular}{l|ccc|ccc}
			\hline
			& \multicolumn{3}{c}{Pearson Correlation ($\uparrow$)} & \multicolumn{3}{|c}{Eigenvector Similarity($\downarrow$)} \\
			\hline
			Methodology   & bn           & uk           & ta           & bn            & uk            & ta      \\
			\hline
			\OurMODEL~($\mathcal{L}_2$)             & 0.47     & 0.36     & 0.42   & 35.55   & 30.64   & 69.72    \\
			\OurMODEL~($\mathcal{L}_{proc_{init}}$) & 0.47     & 0.36     & 0.43   & \textbf{31.23}   & \textbf{10.92}   & \textbf{45.56}    \\
			\OurMODEL~($\mathcal{L}_{proc}$)        & \textbf{0.49}     & \textbf{0.37}     & \textbf{0.44}   & 32.16   & 29.53   & 62.81    \\
			\hline
	\end{tabular}}
	\caption{Analysis of different isometry metrics for~\OurMODEL.}
	\vspace{-3.7ex}
	\label{tab:isometric}
\end{table}

\vspace{-1ex}
\paragraph{Pearson's Correlation.}
We compute Pearson's correlation between the cosine similarities 
of the seed translation pairs as an indicator of the relative 
isomorphism of corresponding spaces. We expect our P@1 results
to correlate positively ($\uparrow$) with Pearson's correlation.

We compute Pearson's correlation over first 1000 translation 
seed pairs. Corresponding results are shown in the first
half of Table~\ref{tab:isometric}. We boldface the best scores.
These results show that for all languages, Pearson's correlation 
for the model~\OurMODEL~($\mathcal{L}_{proc}$) is slightly higher 
compared to other models. 
Although these results are aligned with our findings in Table
~\ref{tab:isometric}, however, one noteworthy observation is that 
Pearson's correlation is not a true indicator of the 
relative performance improvement across different isomorphism losses.

\vspace{-1ex}
\paragraph{Eigenvector Similarity.}
In order to compute the eigenvector similarity of two spaces, 
we compute the Laplacian spectra of the corresponding k-nearest
neighbor graphs. This setting is similar to~
\citet{2018_sogaard}, and is summarized as follows.
For seed pairs construct unweighted graphs followed by
computing the graph Laplacians.
Later, compute the eigenvalues of the graph Laplacians and 
retain the first $k$ eigenvalues summing to less than 90\% of 
the total sum of eigenvalues. Finally, we compute the eigenvector 
similarity as the sum of squared differences between partial 
spectra. The graphs with similar eigenvalue spectra are 
supposed to have similar structures (a measure of 
relative isomorphism).

We expect our eigenvector similarity results to correlate 
negatively ($\downarrow$) with P@1.
The experimental results are shown in the right half of 
Table~\ref{tab:isometric}, with the best scores boldfaced. 
These results show that the eigenvector similarity scores 
for the model~\OurMODEL~($\mathcal{L}_{proc_{init}}$) 
are better than the other two models. This is in contrast 
to our findings in Table~\ref{tab:res1}, where~\OurMODEL~
($\mathcal{L}_{proc}$) shows relatively better performance.

Generally speaking, the results of the isometric metrics 
do not truly correlate with the P@1. These findings are 
aligned with earlier studies by~\citet{2022_isovec} that 
also emphasized the need for better metrics to compute the 
relative isomorphism of the embedding spaces.

\eat{
	This setting is similar to~\citet{2018_sogaard_limit}, and 
	is summarized as follows: 
	for seeds pairs $\{(u_0,v_0), (u_1,v_1),...(u_s,v_s)\}$
	construct unweighted graphs $G_{u}$, $G_{v}$ followed by
	computing the graph laplacians $L_{G} = D_{G} - A_{G}$.
	Later, compute the eigenvalues of the graphs $L_{G_u}$,
	$L_{G_v}$ and retain first $k$ eigenvalues summing to 
	less than 90\% of total sum of eigenvalues for $L_{G_u}$.
	Finally, we compute the eigenvalue similarity as sum of 
	squared differences between partial spectra.}

\subsection{Error Analyses}
We also analyze the error cases of \OurMODEL~in order to 
know the limitations of the model and room for future 
improvement. 
Note, similar to section~\ref{dis:att_conv}, we only perform 
the error analyses for the Ukrainian language and Procrustes loss
($\mathcal{L}_{proc}$).
All experiments were performed with the help of linguistic experts.
We separately analyze the errors for the variants 
of~\OurMODEL~with and without attentive graph convolutions 
(i.e., \OurMODEL; \OurMODEL~w/o G-Conv) in order to quantify 
the reduction in error attributable to the attentive graph 
convolutions.

\begin{table}[b]
	\centering
	\resizebox{0.75\columnwidth}{!}{
		\begin{tabular}{l|cc}
			\hline
			& Type-a & Type-b \\
			\hline
			GRI w/o G-Conv($\mathcal{L}_{proc}$)   & 21.3\% & 6.5\%  \\
			GRI ($\mathcal{L}_{proc}$)             & 50.2\% & 16.6\% \\
			\hline
	\end{tabular}}
	\caption{Classification of Error Types}
	\label{tab:error_count}
\end{table}
In order to better understand the errors from semantic
perspective, we categorize the errors into the following 
types:
\paragraph{Type-a:} The predicted target word for P@1 is semantically 
close to the true target word.
\paragraph{Type-b:} The predicted target word is a k-nearest neighbor 
of the true word for k=5.

We limit the error cases to only the above-mentioned simple types 
because these types give a rough picture of the relative isomorphism 
of the different spaces from the semantic perspective. The 
percentage error counts for both models are shown in Table
~\ref{tab:error_count}. 
For the model \OurMODEL~w/o G-Conv($\mathcal{L}_{proc}$),
21.3\% errors fall in error Type-a,  and 6.5\% errors 
belong to error Type-b.
For the model~\OurMODEL ($\mathcal{L}_{proc}$),
50.2\% errors fall in Type-a, and 16.6\% errors 
belong to Type-b.
As expected the variant of \OurMODEL~with graph convolutions 
shows a higher percentage for both categories, i.e., Type-a and Type-b.
\begin{table}[t]
	\centering
	\resizebox{0.92\columnwidth}{!}{
		\begin{tabular}{ccc|ccc}
			\hline
			\multicolumn{3}{c}{\OurMODEL~($\mathcal{L}_{proc}$)}                & \multicolumn{3}{|c}{\OurMODEL~w/o G-Conv($\mathcal{L}_{proc}$)} \\
			\hline
			\multicolumn{1}{c}{source} & \multicolumn{1}{c}{target} & \multicolumn{1}{c}{target$^{'}$} & 
			\multicolumn{1}{|c}{source} & \multicolumn{1}{c}{target} & \multicolumn{1}{c}{target$^{'}$} \\
			\hline
			\textcyrillic{пошта}       & mail     & mailing   & \textcyrillic{зникли}     & gone       & shattered   \\
			\textcyrillic{спільний}    & shared   & sharing   & \textcyrillic{олія}       & oil        & 60g   \\
			\textcyrillic{вікно}       & window   & windows   & \textcyrillic{банки}      & cans       & merchants   \\
			\textcyrillic{внз}         & college  & teaching  & \textcyrillic{ніс}        & nose       & rubbing  \\
			\textcyrillic{йшов}        & walked   & went      & \textcyrillic{заміна}     & replacing  & overpriced   \\
			\textcyrillic{підручників} & manuals  & templates & \textcyrillic{вулкан}     & volcano    & 100mph    \\
			\textcyrillic{реформа}     & reform   & reforms   & \textcyrillic{ріст}       & growth     & decline   \\   
			\hline
	\end{tabular}}
	\caption{Example error cases for the Ukrainian vs English language for the 
		models: \OurMODEL~($\mathcal{L}_{proc}$); \OurMODEL~w/o G-Conv($\mathcal{L}_{proc}$).
		For each model, the first column (source) corresponds to the Ukrainian words, 
		the second column (target) represents the true target word, third column 
		(target$^{'}$) represents the model predictions for target words.}
	\vspace{-3.7ex}
	\label{tab:error_cases}
\end{table}
These numbers clearly indicate that the attentive graph convolutions
were not only able to correct a major portion of errors made by 
(\OurMODEL~w/o G-Conv), but also the errors made by the model~\OurMODEL~
are either highly semantically related to the target words or a nearest
neighbor of the target word. 

In order to gain further insight, we manually look at the error
cases. For both models, a few examples are shown in Table~
\ref{tab:error_cases}. The majority of the predictions made 
by~\OurMODEL~are indeed correct and closely related to the 
true target words. For example, it predicts \{``mailing", 
``sharing", ``windows"\}
in place of \{``mail", ``shared", ``window"\} respectively.
These results clearly indicate that the current performance of
\OurMODEL~is under-reported and there is a need for better 
quantification measures (other than P@1) in order to compute 
and/or report the true potential of the model.

Overall error analyses show the significance of the 
using attentive graph convolutions to incorporate the 
lexical variations of semantically related tokens in 
order to control the relative isomorphism and perform 
BLI in performance-enhanced way.

\section{Conclusion and Future Directions}
In this paper, we propose Graph-based Relative 
Isomorphism (\OurMODEL) to incorporate the 
impact of lexical variations of semantically 
related tokens in order to control the relative
isomorphism of cross-lingual embeddings.
\OurMODEL~uses multiple different isomorphism losses
(guided by the attentive graph convolutions) along 
with the distributional loss to perform BLI in a
performance-enhanced fashion.
Experimental evaluation shows that \OurMODEL~
outperforms the existing research on BLI by a 
significant margin.
Some probable future directions include:
(i) extending the concepts learned in this research 
to contextualized embeddings, and
(ii) augmenting the~\OurMODEL~to focus more on 
preserving lexico-semantic relations.

\vspace{-1ex}
\section*{Limitations}
\vspace{-1ex}
Some of the core limitations of the proposed approach are as follows:
(i) current formulation of \OurMODEL~is not defined and/or implemented 
for deep contextualized embeddings which are more prevalent and a better
alternate to the distributional embeddings,
(ii) existing limitations of the distributional embeddings are 
inherited in the model, which limits the end-performance 
of~\OurMODEL. For example, as pointed out by~\citet{2019_ANTSYN}
the distributional embedding space tends to inter-mix different
lexico-semantic relations, and yield a poor performance on
a specific task. This phenomenon has a direct impact on~\OurMODEL~
especially on controlling the relative isomorphism of highly 
interlinked relation pairs, e.g., Antonyms vs Synonyms.

\paragraph*{Acknowledgements.}%
Di Wang, Yan Hu and Muhammad Asif Ali are supported in part by the baseline funding BAS/1/1689-01-01, funding from the CRG grand URF/1/4663-01-01, FCC/1/1976-49-01 from CBRC and funding from the AI Initiative REI/1/4811-10-01 of King Abdullah University of Science and Technology (KAUST). Di Wang is also supported by the funding of the SDAIA-KAUST Center of Excellence in Data Science and Artificial Intelligence (SDAIA-KAUST AI).

\bibliography{anthology,custom}
\bibliographystyle{acl_natbib}


\end{document}